# Improving the robustness and accuracy of biomedical language models through adversarial training

Milad Moradi[*]

Institute for Artificial Intelligence, Center for Medical Statistics, Informatics, and Intelligent Systems, Medical University of Vienna, Vienna, Austria

milad.moradivastegani@meduniwien.ac.at

Matthias Samwald

Institute for Artificial Intelligence, Center for Medical Statistics, Informatics, and Intelligent Systems, Medical University of Vienna, Vienna, Austria

matthias.samwald@meduniwien.ac.at

---

[*] Corresponding author. **Postal address:** Institute for Artificial Intelligence, Währinger Straße 25a, 1090 Vienna, Austria. **Telephone number:** 0043-1-40160-36313

## Abstract

Deep transformer neural network models have improved the predictive accuracy of intelligent text processing systems in the biomedical domain. They have obtained state-of-the-art performance scores on a wide variety of biomedical and clinical Natural Language Processing (NLP) benchmarks. However, the robustness and reliability of these models has been less explored so far. Neural NLP models can be easily fooled by adversarial samples, i.e. minor changes to input that preserve the meaning and understandability of the text but force the NLP system to make erroneous decisions. This raises serious concerns about the security and trust-worthiness of biomedical NLP systems, especially when they are intended to be deployed in real-world use cases. We investigated the robustness of several transformer neural language models, i.e. BioBERT, SciBERT, BioMed-RoBERTa, and Bio-ClinicalBERT, on a wide range of biomedical and clinical text processing tasks. We implemented various adversarial attack methods to test the NLP systems in different attack scenarios. Experimental results showed that the biomedical NLP models are sensitive to adversarial samples; their performance dropped in average by 21 and 18.9 absolute percent on character-level and word-level adversarial noise, respectively. Conducting extensive adversarial training experiments, we fine-tuned the NLP models on a mixture of clean samples and adversarial inputs. Results showed that adversarial training is an effective defense mechanism against adversarial noise; the models' robustness improved in average by 11.3 absolute percent. In addition, the models' performance on clean data increased in average by 2.4 absolute present, demonstrating that adversarial training can boost generalization abilities of biomedical NLP systems. This study takes an important step towards revealing vulnerabilities of deep neural language models in biomedical NLP applications. It also provides practical and effective strategies to develop secure, trust-worthy, and accurate intelligent text processing systems in the biomedical domain.

# 1. Introduction

Recent years have witnessed the widespread adoption of Deep Neural Networks (DNNs) for developing intelligent biomedical text processing systems. Vast amounts of freely available biomedical texts, novel DNN architectures, and efficient unsupervised pretraining methods have led to significant breakthroughs in the development of high-performance biomedical language models that effectively encode lexical, syntactic, and semantic regularities of biomedical text. By utilizing these language models as feature extractors or fine-tuning them on task-specific datasets, significant improvements have been achieved on several biomedical NLP tasks such as named entity recognition, relation extraction, text classification, question answering, and summarization [1-7]. However, there are still serious concerns regarding the robustness of neural language models to small changes in input that can be easily handled by humans but cause NLP systems to fail [8]. This issue may be even more critical in the biomedical domain, where wrong decisions by an AI system may have significantly negative consequences.

Recent studies have shown that open-domain NLP models are sensitive to adversarial samples, i.e. small changes to an input such that the modified text can be easily understood by humans and the original meaning is still preserved, but the NLP system is fooled and makes a wrong decision [8-10]. However, no study has so far explored the robustness of biomedical NLP models, hence, knowledge about vulnerabilities of biomedical text processing systems are still limited.

In this paper, we investigate the robustness of high-performance biomedical NLP models against adversarial examples. We implemented various textual adversarial attack methods that produce a wide range of adversarial examples in different attack scenarios. We used the adversarial samples to evaluate the robustness of NLP systems that utilize variations of a state-of-the-art deep neural network transformer model, i.e. Bidirectional Encoder Representations from Transformers (BERT) [11]. The NLP models were already pre-trained on biomedical and clinical text corpora; we fine-tuned and evaluated the models on various biomedical/clinical text processing tasks in adversarial attack scenarios. Evaluations revealed that the biomedical NLP models are not resilient to adversarial samples.

We also performed adversarial training as a defence technique to improve the robustness and security of the NLP models. We fine-tuned the models from scratch on an updated training set that was a mixture of the original training samples and adversarial samples. Experimental results demonstrated that adversarial training can effectively help improve the robustness of NLP systems against adversarial noise in input. Furthermore, experimental results revealed that in addition to serving as a defence mechanism, adversarial training can also act as a regularization technique to reduce overfitting and improve the generalization abilities of the biomedical NLP models. After fine-tuning on adversarial examples, the NLP systems not only performed better on adversarial test data, but also reached new state-of-the-art performance scores on clean (non-adversarial) test samples. This study takes an important step towards enhancing the performance, robustness, and security of biomedical NLP models that will lead to deploying trust-worthy intelligent text processing systems.

The main contributions of our paper are as follows:

- We implemented various textual adversarial attack methods that generate different types of adversarial samples, e.g. white-box, black-box, character-level, and word-level, in the biomedical domain.
- Extensive evaluations on a wide range of biomedical/clinical text processing tasks, we investigated the robustness of several neural biomedical NLP models to adversarial samples and revealed their vulnerabilities.
- Utilizing adversarial samples to conduct adversarial training as a defence technique, we improved the robustness of biomedical NLP models against adversarial noise.
- Utilizing adversarial training as a regularization and data augmentation technique, we enhanced the generalization abilities of biomedical NLP models. This led to new state-of-the-art performance scores on various biomedical/clinical benchmark datasets.

## 2. Related work

Initial work on adversarial attack to deep learning models was devoted to evaluating the robustness of image classifiers against small unperceivable changes in input [12]. However, developing adversarial attack methods to explain oversensitivity points and reveal vulnerabilities of deep neural networks was popularized, and was made simpler and more practical by the Fast Gradient Sign Method (FGSM) proposed by Goodfellow et al. [13]. Adversarial attack methods developed for image data cannot be directly applied to textual data, because images are intrinsically continuous (i.e. pixel values) but texts have a discrete nature. Jia et al. [14] conducted the first study on adversarial attack to textual deep neural networks. They proposed an adversarial evaluation scheme that helps explore language understanding abilities of reading comprehension systems. The idea was to insert sentences to the end of paragraphs in question answering tasks such that the correct answer and human understanding do not change, but NLP systems are distracted. Since then, a wide range of adversarial attack methods targeting different NLP tasks have been proposed in various attack scenarios [8].

Textual adversarial attack methods can be divided into two groups, i.e. white-box and black-box, with respect to the attack method's knowledge of the underlying NLP system. A white-box attack method has full or partial access to information about the NLP model's architecture, loss function, activation functions, parameters, or training data [10, 15, 16]. However, the attack method has usually no access to this information in realistic scenarios. A black-box attacker may only know about probability scores or the final decision made by the NLP model [14, 17-19]. Based on the attack granularity, adversarial samples can be character-level, word-level, or sentences-level [8]. Adversarial attacks may be targeted or untargeted in terms of the attack goal. A targeted attack aims at changing the NLP model's output to a specific text, class label, or value; while the goal of an untargeted attack is to change the output, no matter what the new output is. Adversarial attacks may focus on particular NLP tasks such as text classification [15, 20], malware detection [21, 22], machine translation [16, 23], machine reading comprehension [14, 17], question answering [24], sentiment analysis [25, 26], and textual entailment [27]. In this study, we implemented and utilized various adversarial attack methods, in both white-box and block-box scenarios, with character-level and word-level untargeted attacks, on several biomedical/clinical NLP tasks e.g. text classification, semantic similarity estimation, textual inference, question answering, and relation classification.

A main difference between adversarial attacks to image and textual data is that images are perturbed by injecting small changes to pixel values, which are barely perceptible to humans, while small perturbations to a text can be easily perceived. However, it has been proven that humans can overcome typos, tolerate misspellings, comprehend noisy texts, and deal with synonym words when reading texts [18, 28, 29]. Adversarial attack methods have adopted a wide range of strategies to ensure the perturbed text preserves the same meaning as the original text, it is syntactically correct, and it is still understandable to humans. Common measures used by adversarial attack methods to control perturbations include constraining $L_0$, $L_1$, $L_2$, or $L_\infty$ norms in the continuous vector space [15, 16, 21, 22], allowing limited number of changes [23, 24], measuring semantic equivalency by Euclidean distance or Cosine similarity [26, 30], enforcing grammar and syntax constraints [20, 27], and using perplexity to assess the validity of perturbed text [31]. The adversarial sample generation methods utilized in our experiments employ various measures from the above list to control semantic preservation, syntax and grammar correctness of perturbed texts. We also customized the majority of the attack methods, in order to fit them to biomedical text processing.

Evaluating the robustness of neural networks is not the only purpose of generating adversarial samples. Adversarial training has been shown to be an effective defense strategy to improve the robustness of deep neural networks [13]. Miyato et al. employed adversarial training for the first time on textual deep neural networks [32]. Since then, some studies have explored adversarial training as a robustness boosting and regularization strategy for different NLP tasks such as text classification, reading comprehension, and textual entailment [14, 17, 33, 34]. Our work is the first study that investigates the utility of adversarial training for improving the robustness and accuracy of NLP systems in the biomedical domain.

So far, few studies have addressed textual adversarial sample generation or defending against adversarial attacks to biomedical NLP systems. Previous work covered a limited range of adversarial attack scenarios on few clinical NLP tasks. Sun et al. [35] developed an adversarial attack framework for Long Short-Term Memory (LSTM) networks aiming at identifying susceptible events and measurements in medical records. Wang et al. [36] proposed multi-modal feature consistency check that is a defense method against adversarial attack to clinical summaries of electronic health records. There seems to be much more room for further research in textual adversarial attack and defense in the biomedical domain, which has been remained unexplored by previous work. This paper provides the first extensive study in the biomedical domain that explores various transformer language models in different adversarial attack scenarios on a wide variety of NLP tasks.

## 3. Biomedical NLP tasks

We conducted experiments on five biomedical/clinical text processing datasets covering various NLP tasks. Table 1 presents main statistics of the datasets, as well as state-of-the-art performance scores reported by previous work. In the following, we give a brief description of every task.

**BioText** [37] is a dataset of more than 3.5K text snippets from Medline abstracts annotated for a relation classification task. Eight class labels specify the type of relationship between disease and treatment entities mentioned in every sample.

**MedNLI** [38] contains more than 14K pairs of medical sentences annotated for a textual inference task. Depending on the semantic relation between two sentences, every pair takes one of the labels '*entailment*', '*contradiction*', or '*neutral*'.

**MedSTS** [39] is a dataset of more than 1K pairs of sentences from the Mayo Clinic clinical text database annotated for a semantic similarity task. Every pair takes a similarity value between zero and five, such that five refers to semantic equivalence, while zero signifies complete semantic independence.

**PubMed-RCT** [40] is a text classification dataset containing 200K PubMed abstracts. Every sentence is classified into one of five classes '*background*', '*objective*', '*method*', '*result*', or '*conclusion*' with respect to the role of sentence in the respective abstract.

**PubMed-QA** [41] is a question answering dataset that contains more than 10K research questions, along with their short and long answers extracted from PubMed abstracts.

**Table 1.** Main statistics of the biomedical/clinical NLP datasets. State-of-the-art (SOTA) performance scores reported by previous work are also presented.

| Dataset | Task | Evaluation measure | Number of samples | | | SOTA |
|---|---|---|---|---|---|---|
| | | | **Train** | **Dev** | **Test** | |
| BioText | Relation classification | Micro-F1 | 2,919 | 368 | 368 | 91.7 [42] |
| MedNLI | Textual inference | Accuracy | 11,232 | 1,395 | 1,422 | 88.5 [7] |
| MedSTS | Semantic similarity | Pearson | 675 | 75 | 318 | 84.8 [1] |
| PubMed-RCT | Sentence classification | Micro-F1 | 190K | 5K | 5K | 91.7 [43] |
| PubMed-QA | Question answering | Accuracy | 8000 | 1000 | 1000 | 76.4 [44] |

## 4. Biomedical NLP models

We tested four different biomedical language models that were built on a high-performance transformer deep neural network, i.e. BERT [11]. These models were pre-trained on various biomedical or clinical text corpora and/or using different pre-training strategies. We used the Huggingface Transformers [45] and FARM [46] libraries in python to implement the language models, and fine-tune them on the down-stream tasks. In the following, we give a brief description of the language models.

**BioBERT** [4] utilizes BERT language model [11] with further pre-training on large biomedical text corpora. We used BioBERT-base-v1.1 that was already pre-trained on the PubMed abstracts and PMC full-texts. The bidirectional transformers model utilized by BioBERT-base consists of 12 transformer layers, 768 hidden units in each layer, and 12 attention heads per hidden unit, resulting in a total of 110 million parameters.

**SciBERT** [47]. Unlike BioBERT which was built on the original BERT model (initially pre-trained on two large open-domain corpora, i.e. Wikipedia and Book corpora), SciBERT was pre-trained from scratch on a large multi-domain corpus of scientific text including biomedical publications. The pre-training corpus consisted of more than one million papers from Semantic Scholar [48]. We used the SciBERT-SciVocab-Uncased model as it obtained the best results reported in the respective paper [47].

**BioMed-RoBERTa** [49] was built on RoBERTa-base that optimizes BERT's pre-training approach. The initial model was pre-trained on more data than BERT, with longer sequences and bigger batch sizes. Moreover, a dynamic masking strategy replaced the basic BERT's masking method, also the next sentence prediction objective was excluded from pre-training. BioMed-RoBERTa underwent additional pre-training on a corpus of 2.7 million biomedical scientific papers from Semantic Scholar [48].

**Bio-ClinicalBERT** [50] was initialized from BioBERT-base and further pre-trained on two million clinical notes from the MIMIC-III corpus [51].

# 5. Adversarial attack

In this section, we first give a general formulation of adversarial attack to NLP models, then we describe adversarial attack methods that we utilized in the experiments.

## 5.1. Problem formulation

Given a NLP model $F$ and an input sequence $x=(x_1, x_2, \ldots, x_M)$ such that $x_m$ can be a character, word, or sentence, an attack model $A$ perturbs $x$ and generates $x_{adv}$ to fool the NLP model. The attack model contains two main modules, i.e. Transformation and Search. The Transformation module receives $x$, perturbs it, and produces a set of potential transformations $T=\{t_1, t_2, \ldots, t_N\}$. If a transformation $t_n$ does not satisfy a list of constrains, it is removed from $T$. The Search module successively sends queries to the NLP model and uses a goal function $G$ to select successful adversarial perturbations from the set of transformations. Given the input sequence $x$, a transformation of the input $x_{adv}$, a ground-truth label $y$ such that $F(x)=y$, and a classification result $y'$ such that $F(x_{adv})=y'$, the goal function $G$ determines whether the attack is successful or not. In an untargeted attack scenario, the attack is successful if $y'\neq y$. On the other hand, in a targeted attack scenario, the attack is successful if $y'=y_{target}$, where $y_{target}$ is a prespecified label and $y_{target}\neq y$. In this subsection, we described the adversarial attack to an NLP model on classification task, however, it can be easily generalized to other NLP tasks.

## 5.2. Adversarial attack methods

We implemented four different adversarial attack methods using OpenAttack [52] and TextAttack [53] libraries in python. Adversarial attack strategies are divided into two groups, i.e. *black-box* and *white-box*, based on the attacker's knowledge of the target NLP model. In black-box attack, the attacker has no information about the architecture, parameters, activation functions, loss function, and training data of the NLP model. On the other hand, the attacker has partial or full knowledge of the aforementioned information in a white-box attack scenario. Based on the attack granularity, attackers can generate *character-level* or *word-level* adversarial attacks. Table 2 summarizes the properties of attack methods utilized in our experiments. We give a description of every attack method in the following.

**Table 2.** Properties of the adversarial attack methods regarding knowledge of the target NLP model, attack granularity, and attack target.

| Attack method | White-box | Black-box | Char-level | Word-level | Untargeted | Targeted |
|---|---|---|---|---|---|---|
| HotFlip | ✓ | - | ✓ | ✓ | ✓ | - |
| DeepWordBug | - | ✓ | ✓ | - | ✓ | - |
| TextBugger | ✓ | ✓ | ✓ | ✓ | ✓ | - |
| TextFooler | - | ✓ | - | ✓ | ✓ | - |

**HotFlip** [10] is a white-box attack method that mostly relies on atomic flip operations, i.e. changing one token (i.e. a character or a word) to another based on directional derivatives of the target NLP model with respect to the vector representation of input. It first estimates which individual change to the input will lead to the highest loss, then using a beam search strategy, it searches for an optimal set of perturbations that work well together to fool the NLP model. Four semantic-preserving constraints are applied to word-level swap in order to minimize the chance of changing the meaning of text. Theses constraints are: 1) the Cosine similarity between the embedding of words must be higher than a threshold, 2) the two words must have the same part-of-speech, 3) stop-words must not be replaced, and 4) words with the same lexeme must not replace each other.

**DeepWordBug** [25] is a black-box adversarial attacker that uses four scoring functions, i.e. *Replace-1 Score*, *Temporal Head Score*, *Temporal Tail Score*, and *Combined Score*, in order to identify important tokens in the input. It then applies modifications to the important tokens to enforce the NLP model to make incorrect decisions. *Replace-1 Score* measures the effect of replacing a token $x_m$ with $x'_m$ to estimate the importance of $x_m$. *Temporal Head Score* measures the importance of $x_m$ as the difference between the NLP model's prediction score as it reads up to $x_m$ and the prediction score as it reads up to $x_{m-1}$. *Temporal Tail Score* computes the importance of $x_m$ as the difference between the NLP model's prediction score as it reads the input sequence starting from $x_m$ and the prediction score as it reads starting from $x_{m+1}$. *Combined Score* measures the importance of a token as the weighted sum of *Temporal Head Score* and *Temporal Tail Score*. Transformation functions then apply limited number of character-level changes such as swapping, substitution, deletion, and insertion to the most important tokens.

**TextBugger** [26] can operate in both white-box and black-box attack scenarios. In the white-box setting, the attacker first computes the Jacobian matrix of the NLP model for the input text, using the gradients of the model, in order to identify important words. It then generates five types of bugs, i.e. changes to important words, and chooses the bug that leads to the highest change in the confidence score. The attack method generates character-level adversaries by injecting misspellings to the input. The character-level bugs include insertion, deletion, swap, and substitute. The attacker also produces word-level adversaries by replacing words with their top $K$ nearest neighbours in a context-aware word vector space. For this purpose, we used a GloVe [54] word vector representation model pretrained on the PubMed Central open-access biomedical text corpus[1]. In the black-box setting, TextBugger first uses a scoring function to find words that have the most impact on the NLP model's outcome. The effect of a word on the result is measured by removing that word from the input and computing the difference between the model's confidence score before and after the removal. The bug generation process in the black-box setting is done in the same way as in the white-box setting.

**TextFooler** [55] is a black-box attack method consisting of two main steps: 1) word importance ranking, and 2) word transformation. In the first step, the attacker computes the importance of token $x_m$ by measuring its impact on the NLP model's confidence score for $y$ and $y'$ when $x_m$ appears and does not appear in the input sequence. Input tokens are then ranked based on their importance score. Stop-words are filtered out in the ranking step, in order to avoid grammar destruction. In the second step, given a word $x_m$ with a high importance score, a set of possible replacements is created by identifying top $N$ closest synonyms based on the Cosine similarity between $x_m$ and other words. For this purpose, we used the GloVe word vector representation model pre-trained on the PubMed Central biomedical text corpus. In order to make sure the grammar of the text is not destructed, only those words whose part-of-speech categories are the same as $x_m$ are kept in the set of candidate replacements. At the end of this step, there is a set of candidates $C=\{C_1, C_2, \ldots, C_J\}$ containing the closest synonyms that share the same part-of-speech with $x_m$. Finally, $J$ candidate adversarial samples are generated by replacing $x_m$ with every $C_j$ in $C$. The semantic similarity between every candidate adversarial sample $x_{adv}$ and $x$ is estimated using a BioBERT model trained for semantic similarity estimation, and $x_{adv}$ is added to the list of final adversarial samples if the semantic similarity is not lower than a threshold $\epsilon$.

---

[1] https://www.ncbi.nlm.nih.gov/pmc/tools/openftlist/

## 6. Results and discussion

We first fine-tuned the biomedical NLP models on every dataset described in Section 3. We then tested the fine-tuned models on the test sets to examine how well the models perform on original test data. Afterwards, we evaluated the performance of NLP models on adversarial samples generated by the attack methods. Finally, we trained the NLP models on adversarial samples to improve the robustness against adversarial attack and boost the performance on clean (unperturbed) samples.

### 6.1. Robustness to adversarial samples

Table 3 and Table 4 present the performance scores obtained by the biomedical NLP models on character-level and word-level adversarial samples, respectively. The performance scores on the original (non-adversarial) test samples are also reported in these two tables. Moreover, the tables give the performance decline on adversarial samples in comparison with the scores obtained on the original test data.

As the results show, the NLP models are not robust against adversarial samples. On **character-level** adversarial noise, the average drop in the performance score over all the NLP models is −21.0 on relation classification (BioText), −18.9 on textual inference (MedNLI), −19.8 on semantic similarity estimation (MedSTS), −25.2 on sentence classification (PubMed-RCT), and −20.1 on question answering (PubMed-QA). On **word-level** adversarial noise, the average drop in the performance score over all the NLP models is −18.1 on relation classification, −17.9 on textual inference, −17.8 on semantic similarity estimation, −22.8 on sentence classification, and −17.9 on question answering. The text classification task suffered from the highest drop in performance score. This may suggest that adversarial attack methods can fool NLP models on simpler tasks more easily than tasks with higher complexities. On the other hand, more complex tasks such as textual inference and question answering were less affected by adversarial noise.

In general, the NLP models showed to be more sensitive to character-level adversarial noise. The reason may be that character-level perturbations in the input are mapped to word-piece embeddings that barely appeared in the training data, therefore, no proper representation could be learned for that specific perturbed input during pre-training.

**Table 3.** Performance scores obtained by the biomedical NLP models on the original test samples and **character-level** adversarial samples generated by the attack methods. The scores are reported in percent. The performance decline on adversarial samples is given in parenthesis and is reported in absolute percent. The highest and lowest scores on each task are shown in bold and underlined face, respectively.

| Task | Original/Adversarial | Biomedical NLP models | | | |
|---|---|---|---|---|---|
| | | **BioBERT** | **SciBERT** | **BioMed-RoBERTa** | **Bio-ClinicalBERT** |
| BioText | Original test set | 91.3 | 90.5 | 92.4 | 89.0 |
| | HotFlip | 68.7 (−22.6) | 67.2 (−23.3) | 71.1 (−21.3) | 67.7 (−21.3) |
| | DeepWordBug | 70.3 (−21.0) | 69.0 (−21.5) | 71.8 (−20.6) | 69.5 (−19.5) |
| | TextBugger Black-box | 69.1 (−22.2) | <u>65.9 (−24.6)</u> | 69.5 (−22.9) | 66.8 (−22.2) |
| | TextBugger White-box | 70.5 (−20.8) | 71.4 (−19.1) | **74.9 (−17.5)** | 73.2 (−15.8) |
| MedNLI | Original test set | 85.6 | 84.9 | 87.7 | 86.5 |
| | HotFlip | <u>64.0 (−21.6)</u> | 64.6 (−20.3) | 65.9 (−21.8) | 67.2 (−19.3) |
| | DeepWordBug | 65.8 (−19.8) | 66.9 (−18.0) | 65.8 (−21.9) | **71.8 (−14.7)** |
| | TextBugger Black-box | 67.5 (−18.1) | 66.1 (−18.8) | 68.2 (−19.5) | 70.6 (−15.9) |
| | TextBugger White-box | 66.7 (−18.9) | 67.3 (−17.6) | 68.5 (−19.2) | 68.5 (−18.0) |
| MedSTS | Original test set | 86.0 | 86.8 | 88.1 | 88.3 |
| | HotFlip | 65.9 (−20.1) | <u>65.5 (−21.3)</u> | 67.7 (−20.4) | 70.2 (−18.1) |
| | DeepWordBug | 67.3 (−18.7) | 66.7 (−20.1) | 67.4 (−20.7) | 69.5 (−18.8) |
| | TextBugger Black-box | 66.9 (−19.1) | <u>65.5 (−21.3)</u> | 66.8 (−21.3) | 68.6 (−19.7) |
| | TextBugger White-box | 67.0 (−19.0) | 66.1 (−20.7) | 68.2 (−19.9) | **70.5 (−17.8)** |
| PubMed-RCT | Original test set | 92.4 | 91.3 | 92.8 | 91.6 |
| | HotFlip | 67.3 (−25.1) | 65.8 (−25.5) | 67.7 (−25.1) | <u>64.1 (−27.5)</u> |
| | DeepWordBug | 68.5 (−23.9) | 68.5 (−22.8) | **70.2 (−22.6)** | 66.0 (−25.6) |
| | TextBugger Black-box | 64.9 (−27.5) | 66.0 (−25.3) | 65.5 (−27.3) | 65.5 (−26.1) |
| | TextBugger White-box | 68.0 (−24.4) | 66.0 (−25.3) | 66.4 (−26.4) | 67.8 (−23.8) |
| PubMed-QA | Original test set | 81.7 | 82.1 | 83.5 | 82.7 |
| | HotFlip | 62.7 (−19.0) | 63.4 (−18.7) | 62.9 (−20.6) | 61.5 (−21.2) |
| | DeepWordBug | **65.1 (−16.6)** | 62.6 (−19.5) | 62.8 (−20.7) | 62.2 (−20.5) |
| | TextBugger Black-box | 62.0 (−19.7) | 63.3 (−18.8) | 61.9 (−21.6) | <u>59.5 (−23.2)</u> |
| | TextBugger White-box | 61.5 (−20.2) | 63.3 (−18.8) | 63.4 (−20.1) | 60.5 (−22.2) |
| Mean (all tasks-original test samples) | | 87.4 | 87.1 | 88.9 | 87.6 |
| Mean (all tasks-adversarial samples) | | 66.5 (−20.9) | 66.1 (−21.0) | 67.3 (−21.6) | 67.1 (−20.5) |

**Table 4.** Performance scores obtained by the biomedical NLP models on the original test samples and **word-level** adversarial samples generated by the attack methods. The scores are reported in percent. The performance decline on adversarial samples is given in parenthesis and is reported in absolute percent. The highest and lowest scores on each task are shown in bold and underlined face, respectively.

| Task | Original/Adversarial | Biomedical NLP models | | | |
|---|---|---|---|---|---|
| | | **BioBERT** | **SciBERT** | **BioMed-RoBERTa** | **Bio-ClinicalBERT** |
| BioText | Original test set | 91.3 | 90.5 | 92.4 | 89.0 |
| | HotFlip | 71.1 (−20.2) | 69.3 (−21.2) | 73.2 (−19.2) | 70.5 (−18.5) |
| | TextBugger Black-box | 72.2 (−19.1) | 71.4 (−19.1) | 74.9 (−17.5) | 72.9 (−16.1) |
| | TextBugger White-box | 70.3 (−21.0) | <u>68.8 (−21.7)</u> | 73.1 (−19.3) | 69.2 (−19.8) |
| | TextFooler | 73.6 (−17.7) | 76.1 (−14.4) | **78.7 (−13.7)** | 77.8 (−11.2) |
| MedNLI | Original test set | 85.6 | 84.9 | 87.7 | 86.5 |
| | HotFlip | <u>65.6 (−20.0)</u> | 66.8 (−18.1) | 67.4 (−20.3) | 69.8 (−16.7) |
| | TextBugger Black-box | 67.1 (−18.5) | 68.3 (−16.6) | 67.0 (−20.7) | 70.5 (−16.0) |
| | TextBugger White-box | 66.3 (−19.3) | 67.9 (−17.0) | 68.1 (−19.6) | 69.8 (−16.7) |
| | TextFooler | 68.5 (−17.1) | 69.2 (−15.7) | 69.8 (−17.9) | **70.9 (−15.6)** |
| MedSTS | Original test set | 86.0 | 86.8 | 88.1 | 88.3 |
| | HotFlip | 67.5 (−18.5) | <u>66.9 (−19.9)</u> | 69.2 (−18.9) | 71.3 (−17.0) |
| | TextBugger Black-box | 69.1 (−16.9) | 68.8 (−18.0) | 69.9 (−18.2) | 72.1 (−16.2) |
| | TextBugger White-box | 68.0 (−18.0) | 68.2 (−18.6) | 69.9 (−18.2) | 71.0 (−17.3) |
| | TextFooler | 68.8 (−17.2) | 67.9 (−18.9) | 70.5 (−17.6) | **72.2 (−16.1)** |
| PubMed-RCT | Original test set | 92.4 | 91.3 | 92.8 | 91.6 |
| | HotFlip | 69.7 (−22.7) | 68.8 (−22.5) | 70.9 (−21.9) | <u>66.8 (−24.8)</u> |
| | TextBugger Black-box | **71.8 (−20.6)** | 70.4 (−20.9) | 71.1 (−21.7) | 68.3 (−23.3) |
| | TextBugger White-box | 67.5 (−24.9) | 68.6 (−22.7) | 68.2 (−24.6) | 67.0 (−24.6) |
| | TextFooler | 70.6 (−21.8) | 69.0 (−22.3) | 69.5 (−23.3) | 68.8 (−22.8) |
| PubMed-QA | Original test set | 81.7 | 82.1 | 83.5 | 82.7 |
| | HotFlip | 64.5 (−17.2) | 65.1 (−17.0) | 65.1 (−18.4) | 63.6 (−19.1) |
| | TextBugger Black-box | 66.2 (−15.5) | 65.9 (−16.2) | **66.4 (−17.1)** | 64.3 (−18.4) |
| | TextBugger White-box | 65.0 (−16.7) | 64.2 (−17.9) | 65.5 (−18.0) | <u>61.7 (−21.0)</u> |
| | TextFooler | 64.5 (−17.2) | 63.3 (−18.8) | 65.1 (−18.4) | 62.4 (−20.3) |
| Mean (all tasks-original test samples) | | 87.4 | 87.1 | 88.9 | 87.6 |
| Mean (all tasks-adversarial samples) | | 68.4 (−19.0) | 68.2 (−18.9) | 69.7 (−19.2) | 69.0 (−18.6) |

As the results show, white-box attack methods, i.e. HotFlip and the white-box variation of TextBugger, generally caused larger declines in performance scores compared to the black-box attackers. This demonstrates that having access to the underlying NLP model's gradients or loss function can help the attack methods find those inputs that have the highest impact on changing the output more easily than black-box attackers. Observing the performance scores, it seems that pre-training strategy and datasets used for pre-training are important contributing factors to the NLP system's robustness against adversarial samples. For example, BioMed-RoBERTa obtained the highest after-attack scores on two datasets, i.e. BioText and PubMed-RCT in character-level attack settings, BioText and PubMed-QA in word-level attack settings. What makes it different from the other BERT-family models is the efficient pre-training strategy (i.e. longer sequences, a bigger batch size, and dynamic masking) that this model utilized. Another example, Bio-ClinicalBERT obtained the highest after-attack scores on the two clinical NLP tasks, i.e. MedNLI and MedSTS. Pretraining on in-domain clinical texts helped this model be less sensitive than the other NLP models to adversarial samples.

Figure 1 shows four examples of adversarial samples for which BioBERT failed to produce the correct output. The examples were selected from four different biomedical/clinical NLP tasks. In this figure, example 1 and examples 2 show character-level adversarial samples from the relation classification and semantic similarity tasks, respectively. The NLP model was fooled on these examples, although the perturbed texts are still understandable and convey the same meaning as the original texts. Example 3 and example 4 show word-level adversarial samples from the textual inference and text classification tasks, respectively. In these examples, some words have been replaced with their synonyms, but the meaning has been preserved. However, the NLP model could not make correct decisions on these samples.

## 6.2. Adversarial training

As a mechanism of defense against adversarial noise, NLP models can be trained on adversarial samples in order to enhance their robustness [8, 15, 56]. Moreover, adversarial training can act as a regularization technique to improve the generalization performance of NLP systems [8, 29]. We performed adversarial training by fine-tuning the models from scratch on a mixture of original training samples and adversarial samples produced on the training data, separately for every biomedical NLP task. We then evaluated the performance of NLP models on adversarial test samples to examine the effect of adversarial training on robustness. We also evaluated the models on clean test samples to see if adversarial training can enhance the models' generalization performance. We separately fine-tuned and tested the NLP models on character-level, word-level, and all adversarial samples to examine the effect of different attack granularities on the robustness and performance.

### 6.2.1. Robustness improvement

Table 5 presents performance scores obtained by the biomedical NLP models on adversarial test samples, after fine-tuning the models on adversarial training samples. The absolute performance improvement achieved on adversarial samples, after adversarial training, is also presented in this table. The performance improvement was calculated with respect to the average of performance scores obtained by each NLP model on character-level, word-level, and all adversarial samples reported in Table 3 and Table 4.

**Example 1:**

Relation classification: BioText

**Original sample:** On the 2 days of testing, the child received AEDs and a capsule containing either placebo or methylphenidate.
**Adversarial sample:** On the 2 days of testting, the child recieved AEDs and a capsul containing either plaecbo or methylphenidate.
**Ground-truth label:** *Treatment_Only*
**Predicted label:** *None*

**Example 2:**

Semantic similarity estimation: MedSTS

**Original sample:** He denies any shortness of breath or difficulty breathing. Patient denies any chest pain or shortness of breath.
**Adversarial sample:** He denies an shotrness of breath or diffculty braething. Pattient denies any chest pain or shorttness of breath.
**Ground-truth similarity:** *4.5*
**Estimated similarity:** *3*

**Example 3:**

Textual inference: MedNLI

**Original sample:** EKG exhibited ST elevation inferiorly and Nuclear MIBI showed lateral wall defect. The patient has an acute STEMI.
**Adversarial sample:** EKG showed ST increase inferiorly and Nuclear MIBI showed lateral wall deficiency. The patient has an severe STEMI.
**Ground-truth label:** *Entailment*
**Predicted label:** *Neutral*

**Example 4:**

Text classification: PubMed-RCT

**Original sample:** The best strategy to protect individuals against meningococcal disease is to immunize against multiple serogroups.
**Adversarial sample:** The best approach to preserve people against meningococcal illness is to immunize against several serogroups.
**Ground-truth label:** *Background*
**Predicted label:** *Methods*

**Figure 1.** Four examples of adversarial samples for which BioBERT failed to produce the correct output. The examples were selected from four different biomedical/clinical text processing tasks. The first two examples show character-level adversarial samples, while the last two examples show word-level adversarial samples.

As the results show, the performance scores improved (with respect to Table 3 and Table 4), demonstrating that adversarial training has substantial positive effect on the robustness of the NLP systems. Including only **character-level** adversarial samples in adversarial training, the average absolute improvement over all the NLP models is +12.8 on relation classification (BioText), +10.8 on textual inference (MedNLI), +9.1 on semantic similarity estimation (MedSTS), +14.4 on sentence classification (PubMed-RCT), and +10.6 on question answering (PubMed-QA). Including only **word-level** adversarial samples in adversarial training, the average absolute improvement over all the NLP models is +10.6 on relation classification, +7.8 on textual inference, +6.4 on semantic similarity estimation, +12.2 on sentence classification, and +8.8 on question answering. Including **both attack granularities** in adversarial training, the average absolute improvement over all the NLP models is +14.9 on relation classification, +11.9 on textual inference, +10.7 on semantic similarity estimation, +16.2 on sentence classification, and +12.4 on question answering.

**Table 5.** Performance scores obtained by the biomedical NLP models on adversarial samples, after adversarial training. The NLP models were separately trained on character-level, word-level, and both attack granularities. The NLP models were tested on all character-level and word-level adversarial samples generated by all the attack methods to investigate how adversarial training affects the robustness to adversarial attack. The scores are reported in percent. The improvement on adversarial samples, achieved after adversarial training, is given in parenthesis. The improvement is reported in absolute percent. The highest and lowest scores on each task are shown in bold and underlined face, respectively.

| Task | Adversarial training granularity | Biomedical NLP models | | | |
|------|------|------|------|------|------|
| | | **BioBERT** | **SciBERT** | **BioMed-RoBERTa** | **Bio-ClinicalBERT** |
| BioText | Char-level | 84.1 (+13.4) | 82.9 (+13.1) | 85.0 (+11.7) | 83.9 (+13.0) |
| | Word-level | 82.8 (+12.1) | 80.7 (+10.9) | 82.6 (+09.3) | 81.1 (+10.2) |
| | Char- and Word-level | 86.5 (+15.8) | 84.8 (+15.0) | **87.7 (+14.4)** | 85.4 (+14.5) |
| MedNLI | Char-level | 77.3 (+10.9) | 77.3 (+10.2) | 79.2 (+11.7) | 80.4 (+10.6) |
| | Word-level | 74.0 (+07.6) | 73.5 (+06.4) | 77.3 (+09.8) | 77.3 (+07.5) |
| | Char- and Word-level | 79.3 (+12.9) | 78.6 (+11.5) | 79.5 (+12.0) | **80.9 (+11.1)** |
| MedSTS | Char-level | 77.9 (+10.4) | 75.8 (+08.9) | 77.9 (+09.3) | 78.3 (+07.7) |
| | Word-level | 75.5 (+08.0) | 73.9 (+07.0) | 75.0 (+06.4) | 75.0 (+04.4) |
| | Char- and Word-level | 79.7 (+12.2) | 76.5 (+09.6) | **80.4 (+11.8)** | 79.7 (+09.1) |
| PubMed-RCT | Char-level | 83.2 (+14.7) | 81.1 (+13.3) | 84.5 (+15.9) | 80.4 (+13.7) |
| | Word-level | 80.8 (+12.3) | 79.5 (+11.7) | 81.7 (+13.1) | 78.3 (+11.6) |
| | Char- and Word-level | 85.6 (+17.1) | 83.8 (+16.0) | **86.0 (+17.4)** | 81.1 (+14.4) |
| PubMed-QA | Char-level | 74.0 (+10.1) | 73.8 (+10.0) | 75.7 (+11.6) | 72.9 (+11.0) |
| | Word-level | 72.7 (+08.8) | 71.6 (+07.8) | 73.0 (+08.9) | 71.5 (+09.6) |
| | Char- and Word-level | 76.4 (+12.5) | 75.9 (+12.1) | **77.1 (+13.0)** | 74.0 (+12.1) |
| Mean (all tasks-adversarial samples) | | 79.3 (+11.9) | 78.0 (+10.9) | 80.2 (+11.7) | 78.7 (+10.7) |

The results demonstrate that character-level adversarial training leads to higher improvement in the robustness of the models, compared with word-level adversarial training. However, including both granularities has the highest positive effect on the robustness. In general, the highest improvement was achieved on the text classification task. On the other hand, the semantic similarity, textual inference, and question answering tasks reported the lowest amounts of improvement. This suggests that simpler tasks may benefit from adversarial training more than tasks with higher complexities. However, as we discussed in Section 6.1, fooling NLP models on simpler tasks, e.g. text classification, is easier than on complex tasks such as textual inference and question answering.

Although Table 5 reports a negligible difference in the average performance improvement gained by the NLP systems after adversarial training, BioMed-RoBERTa achieved the highest scores on adversarial samples on four of the five tasks. It also obtained the best performance scores on the original test data, and achieved the highest after-attack scores on two tasks, as presented in Table 3 and Table 4. This implies the high resilience of this NLP model under different training and testing scenarios, which makes it a notable candidate to be deployed as an intelligent biomedical text processing system in real-world use cases.

### 6.2.2. Generalization performance improvement

Table 6 presents performance scores obtained by the biomedical NLP models on normal (non-adversarial) test samples, after fine-tuning the models on adversarial training samples. The absolute performance improvement achieved on original test samples, after adversarial training, is also presented in this table. The performance improvement was calculated with respect to the performance scores obtained by each NLP model on original test samples reported in Table 3 and Table 4.

As the results show, the NLP models surpassed state-of-the-art performance scores reported by previous work. This demonstrates that adversarial training is an effective regularization technique to enhance the generalization abilities of biomedical NLP models. Including only **character-level** adversarial samples in adversarial training, the average absolute improvement over all the NLP models is +1.8 on relation classification (BioText), +1.1 on textual inference (MedNLI), +1.6 on semantic similarity estimation (MedSTS), +0.7 on sentence classification (PubMed-RCT), and +1.2 on question answering (PubMed-QA). Including only **word-level** adversarial samples in adversarial training, the average absolute improvement over all the NLP models is +3.0 on relation classification, +2.9 on textual inference, +3.3 on semantic similarity estimation, +1.9 on sentence classification, and +2.5 on question answering. Including **both attack granularities** in adversarial training, the average absolute improvement over all the NLP models is +3.6 on relation classification, +3.5 on textual inference, +3.9 on semantic similarity estimation, +2.3 on sentence classification, and +3.1 on question answering.

The results demonstrate that including both attack granularities has the highest positive impact on the effectiveness of adversarial training. However, word-level adversarial training leads to higher improvement in the performance of the NLP models on clean test samples, compared with character-level adversarial training. On the other hand, fine-tuning on character-level attacks led to higher robustness improvement, as already discussed in Section 6.2.1. This implies that adversarial training on both character-level and word-level attacks is necessary for enhancing the robustness and generalization performance of biomedical NLP models.

**Table 6.** Performance scores obtained by the biomedical NLP models on original test samples, after adversarial training. The NLP models were separately trained on character-level, word-level, and both attack granularities. The NLP models were tested on the original test sets to investigate how adversarial training affects the performance on normal (non-adversarial) test samples. The scores are reported in percent. The improvement on original test samples, achieved after adversarial training, is given in parenthesis. The improvement is reported in absolute percent. The highest and lowest scores on each task are shown in bold and underlined face, respectively.

| Task | Adversarial training granularity | Biomedical NLP models | | | |
|---|---|---|---|---|---|
| | | **BioBERT** | **SciBERT** | **BioMed-RoBERTa** | **Bio-ClinicalBERT** |
| BioText | Char-level | 92.2 (+00.9) | <u>91.5 (+01.0)</u> | 93.1 (+00.7) | 93.5 (+04.5) |
| | Word-level | 93.5 (+02.2) | 93.5 (+03.0) | 94.1 (+01.7) | 94.1 (+05.1) |
| | Char- and Word-level | 94.1 (+02.8) | 93.5 (+03.0) | **95.3 (+02.9)** | 94.7 (+05.7) |
| MedNLI | Char-level | 86.9 (+01.3) | <u>86.1 (+01.2)</u> | 88.1 (+00.4) | 88.1 (+01.6) |
| | Word-level | 88.5 (+02.9) | 87.7 (+02.8) | 90.3 (+02.6) | 89.8 (+03.3) |
| | Char- and Word-level | 89.2 (+03.6) | 88.5 (+03.6) | 90.3 (+02.6) | **90.8 (+04.3)** |
| MedSTS | Char-level | <u>87.1 (+01.1)</u> | 88.6 (+01.8) | 90.3 (+02.2) | 89.8 (+01.5) |
| | Word-level | 88.6 (+02.6) | 89.8 (+03.0) | 92.0 (+03.9) | 91.9 (+03.6) |
| | Char- and Word-level | 89.5 (+03.5) | 90.3 (+03.5) | **92.5 (+04.4)** | 92.5 (+04.2) |
| PubMed-RCT | Char-level | 92.8 (+00.4) | <u>92.5 (+01.2)</u> | 93.1 (+00.3) | 92.5 (+00.9) |
| | Word-level | 94.1 (+01.7) | 93.7 (+02.4) | 94.0 (+01.2) | 94.0 (+02.4) |
| | Char- and Word-level | 94.6 (+02.2) | 94.0 (+02.4) | **94.9 (+02.1)** | 94.0 (+02.4) |
| PubMed-QA | Char-level | 83.3 (+01.6) | 83.9 (+01.8) | 84.7 (+01.2) | <u>83.1 (+00.4)</u> |
| | Word-level | 84.5 (+02.8) | 84.5 (+02.4) | 86.4 (+02.9) | 84.8 (+02.1) |
| | Char- and Word-level | 85.1 (+03.4) | 85.6 (+03.5) | **87.0 (+03.5)** | 84.8 (+02.1) |
| Mean (all tasks-original test samples) | | 89.6 (+02.2) | 89.6 (+02.4) | 91.1 (+02.2) | 90.5 (+02.9) |

As we discussed in Section 6.2.1, simpler tasks gained higher robustness improvement after adversarial training. However, Table 6 reports higher performance improvement on tasks with higher complexities such as semantic similarity and textual inference. This suggests that complex tasks may need more training samples to achieve higher levels of predictive accuracy; including adversarial samples in the training data can be an effective data augmentation technique to provide biomedical NLP models with additional in-domain knowledge and boost their generalization performance. Although the difference between the average performance improvements gained by the NLP models may not be significant, BioMed-RoBERTa outperforms the other models on four of the five tasks. Regarding its superiority to the other models on the majority of the evaluations, BioMed-RoBERTa has shown to have considerable capabilities in mastering different biomedical text processing tasks and benefitting from adversarial training. We have released the robust BioMed-RoBERTa models on Huggingface's transformers hub. They are publicly available and can be utilized to develop biomedical/clinical text processing systems. Links to the models for every task can be found in Table 7. Supplementary materials and source codes are available at https://github.com/mmoradi-iut/RobustBioNLP.

**Table 7.** Links to the robust BioMed-RoBERTa models for every biomedical/clinical NLP task used in our experiments.

| Model name | Link |
| --- | --- |
| Robust-BioMed-RoBERTa-RelationClassification | https://huggingface.co/mmoradi/Robust-Biomed-RoBERTa-RelationClassification |
| Robust-BioMed-RoBERTa-TextualInference | https://huggingface.co/mmoradi/Robust-Biomed-RxoBERTa-TextualInference |
| Robust-BioMed-RoBERTa-SemanticSimilarity | https://huggingface.co/mmoradi/Robust-Biomed-RoBERTa-SemanticSimilarity |
| Robust-BioMed-RoBERTa-TextClassification | https://huggingface.co/mmoradi/Robust-Biomed-RoBERTa-TextClassification |
| Robust-BioMed-RoBERTa-QuestionAnswering | https://huggingface.co/mmoradi/Robust-Biomed-RoBERTa-QuestionAnswering |

# 7. Conclusion

In this paper, we evaluated the robustness of several biomedical transformer neural language models against adversarial samples on various biomedical/clinical NLP tasks. The experimental results demonstrated that the NLP models are unstable to adversarial noise in the input. This raises serious concerns about the security and trustworthiness of intelligent systems for real-world biomedical text processing applications. We also conducted extensive adversarial training experiments that showed including adversarial examples in training data can be an effective strategy to improve the robustness of biomedical NLP systems. Moreover, the results suggested that adversarial training can act as an efficient data augmentation and regularization technique to enhance the generalization abilities of biomedical language models.

This study revealed vulnerabilities of deep neural NLP models to textual adversaries in the biomedical domain. Evaluating NLP models under adversarial settings can help us identify sensitivity points of intelligent text processing systems in realistic scenarios. As previous work has shown, common performance measures such as accuracy, precision, recall, etc. may not be informative enough to assess the utility of a NLP system in real-world applications [57, 58]. Evaluation benchmarks should also consider security measures such as robustness to adversarial and natural noise. In this way, users of these NLP models (e.g. developers, physicians, clinicians, etc.) will be able to have a more realistic assessment of how accurate and robust these intelligent systems can operate in the real-world. Explainable Artificial Intelligence (XAI) methods have shown to be effective in explaining decision boundaries of NLP systems [42, 59-61]. Future work may include combining XAI methods and textual adversaries to interpret decisions of a black-box biomedical NLP model, explain why and how it fails in different parts of the decision space, and suggest how it can be optimized and improved.